\documentclass[sigconf]{acmart}

\usepackage[utf8]{inputenc} 
\usepackage[T1]{fontenc}    
\usepackage{hyperref}       
\usepackage{url}            
\usepackage{booktabs}       
\usepackage{amsfonts}       
\usepackage{nicefrac}       
\usepackage{microtype}      
\usepackage{xcolor}
\usepackage{graphicx}
\usepackage{wrapfig}
\usepackage{booktabs}       
\usepackage{tabularx}
\usepackage[notheorems]{dgleich-math}
\usepackage[font=small,labelfont=bf]{caption}

\newcolumntype{Y}{>{\raggedright\arraybackslash}X}
\newcolumntype{W}{>{\raggedleft\arraybackslash}X}
\newcolumntype{Z}{>{\centering\arraybackslash}X}
\newcolumntype{H}{>{\setbox0=\hbox\bgroup}c<{\egroup}@{}}
\AtBeginDocument{%
  \providecommand\BibTeX{{%
    \normalfont B\kern-0.5em{\scshape i\kern-0.25em b}\kern-0.8em\TeX}}}

\graphicspath{{figure/}}

\renewcommand\footnotetextcopyrightpermission[1]{} 

\setcopyright{none}
\begin{document}
\fancyhead{}

\title{Zero-Shot Heterogeneous Transfer Learning from Recommender Systems to Cold-Start Search Retrieval}


\author{
  Tao Wu\text{*},
  Ellie Ka-In Chio\text{*},
  Heng-Tze Cheng\text{*}
}

\author{
  Yu Du, Steffen Rendle, Dima Kuzmin, Ritesh Agarwal, Li Zhang, John Anderson, Sarvjeet Singh, Tushar Chandra, Ed H. Chi, Wen Li, Ankit Kumar, Xiang Ma, Alex Soares, Nitin Jindal, Pei Cao}

\affiliation{%
  \institution{Google Inc.$^\dagger$}
}

\thanks{$*$ Equal contribution}
\thanks{$\dagger$ Corresponding author: Tao Wu: \texttt{iotao@google.com}}

\renewcommand{\shortauthors}{Wu, Tao, et al.}



\begin{abstract}
Many recent advances in neural information retrieval models, 
which predict top-$K$ items given a query, learn directly from a large training set of (query, item) pairs.
However, they are often insufficient when there are many previously unseen (query, item) combinations, often referred to as the cold start problem.
Furthermore, the search system can be biased towards items that are frequently shown to a query previously, also known as the ``rich get richer'' (a.k.a. feedback loop) problem.
In light of these problems, we observed that most online content platforms have both a search and a recommender system that, while having heterogeneous input spaces, can be connected through their common output item space and a shared semantic representation.
In this paper, we propose a new Zero-Shot Heterogeneous Transfer Learning framework that transfers learned knowledge from the recommender system component to improve the search component of a content platform.
First, it learns representations of items and their natural-language features by predicting (item, item) correlation graphs derived from the recommender system as an auxiliary task.
Then, the learned representations are transferred to solve the target search retrieval task, performing query-to-item prediction without having seen any (query, item) pairs in training.
We conduct online and offline experiments on
one of the world's largest search and recommender systems from Google, and present the results and lessons learned. 
We demonstrate that the proposed approach can achieve high performance on offline search retrieval tasks,
and more importantly, achieved significant improvements on relevance and user interactions over the highly-optimized production system in online experiments. 
\end{abstract}


\copyrightyear{}
\acmYear{}
\acmConference[]{}{}{}
\acmDOI{}
\acmISBN{}



\maketitle


\section{Introduction}


Most online content platforms, such as music streaming services or e-commerce websites, have systems that return top-$K$ items either given a natural-language query (i.e., a search retrieval system), or given the user context, which can be the user attributes and user's interactions with the platform (i.e., a recommender system). These two systems share the same output item space, but have different input feature spaces.
In this paper, we study how to improve search retrieval by transferring learnings from recommender systems.

Recently neural information retrieval (neural IR) models have been widely applied in search products across many industries~\cite{haldar2019applying,zhai2019learning,ramanath2018towards}. 
Such methods can retrieve and score items that do not share keywords with the query.
However, they usually 
require large amount of (query, item) pairs as training data, usually collected from users' search logs.
However, these types of training data may not be available for many (query, item) combinations. This is often referred to as the \textit{cold start} problem~\cite{lam2008addressing}.
Consider, for example, an online music streaming system, where users mostly listen to music through homepage recommendations or playlists generated by other users. In this case, it is likely that many songs in the system have been listened by users, but not through search requests.
This motivated us to ask: Can we utilize the abundant data from the recommender system to cold-start the search retrieval system of the same content platform?

\begin{figure*}[!ht]
  \begin{center}
    \includegraphics[width=0.85\textwidth]{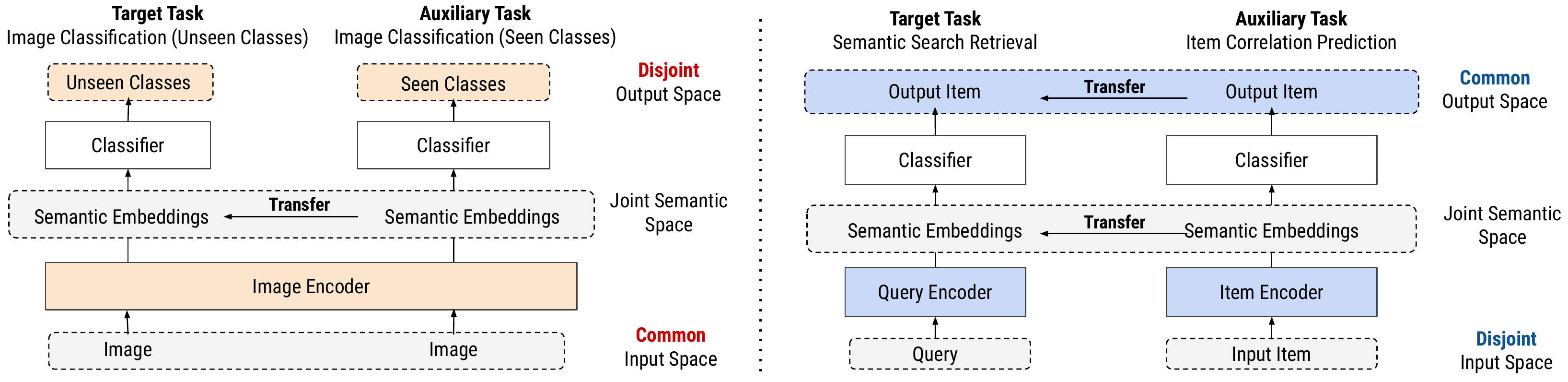}
  \end{center}
  \vspace{-0.1cm}
  \caption{A comparison on the high-level frameworks between the classic zero-shot learning in image classification~\cite{socher2013zero} (on the left) and the zero-shot heterogeneous transfer learning in this paper (on the right).}
  \label{illustration}
\end{figure*}

In this paper, we propose a new Zero-Shot Heterogeneous Transfer Learning framework (ZSL), which does not require any input from (query, item) search data, but learns the query and item representations completely via an auxiliary task derived from the recommender system. There are two advantages: first, this method can cold-start the search system as it does not require search data to train. Second, supervised methods trained on (query, item) can suffer from the potential bias and feedback loop~\cite{sinha2016deconvolving} introduced by the search system.
For large-scale search and recommender systems, there could easily be more than thousands of relevant items matching a query, but users are only able to explore a very limited amount. Therefore, the collected training data are heavily affected by the current search algorithm, exacerbating the "rich get richer" effect~\cite{chen2019top}. Thus, this framework, although motivated by the search retrieval cold-start problem, can also be useful even when adequate search data (query, item) is available.

We assume that (item, item) correlations can be extracted from the recommender system, with text features for each item.
The (item, item) correlations commonly exist~\cite{hu2008collaborative}, such as citation network of a bibliography system or music co-listening pairs of an online music platform, where the text features of items are titles, keywords or descriptions. The auxiliary task of the proposed framework is to predict neighbor items given a seed item, where the learned semantic representations will be transferred to the target task which is to predict items given a query.
We explore two implementations under this framework.
We call the first method Multi-task Item Encoder, where the item and the text features representations are jointly learned by optimizing the two tasks of predicting the item given its text features, and predicting the item given it neighbors. The second approach is Item-to-Text Transformed Encoder, where it only optimizes for a single task of predicting the item given its neighbors, but utilizing the text features to encode the items.
We conduct experiments on one of the world's largest recommender systems from Google. 
Our proposed methods demonstrate promising results on multiple offline retrieval tasks.
In a A/B test live experiment study, where the base is the production search system that is already highly fine tuned, with many components of advanced retrieval techniques (e.g., both term-frequency-based scoring and supervised machine learning algorithms), our proposed method improved multiple evaluation metrics. 
This shows that even for a system with enough search training data available, ensembling our proposed method can successfully introduce new relevant results that are favored by users.

Our contributions in this paper are summarized below:
\begin{enumerate}
    \item To our best knowledge, this is the first work that studies the problem of cold-starting a production-scale search retrieval system from (item, item) correlations in the recommender system from the same online content platform.
    \item We proposed the Zero-Shot Heterogeneous Transfer Learning framework as a solution to cold-start the search retrieval system.
    \item We conduct extensive offline and online experiments on one of the world's largest recommender systems, and find that our proposed method 1) when applied alone, can produce accurate retrieval results; 2) when ensembled with supervised methods, can improve the highly optimized search retrieval system.
    In addition, our findings regarding the effectiveness of our method on broad queries (inferred by query lengths), are valuable insights for practitioners to apply such techniques to real word search systems. 
\end{enumerate}

\section{Related Work}




\textbf{Zero-shot learning and transfer learning.}
For large-scale label domains, it is common to have labels of instances that have never been seen in the training data.
The key idea of Zero-shot learning~\cite{wang2019survey} is to utilize some auxiliary information of the unseen labels, and learn a model to connect the auxiliary information to the input space. By mapping the input feature to the auxiliary information, the zero-shot algorithms are then able to find the corresponding unseen labels. Applications of such idea include object detection of unseen classes~\cite{socher2013zero}, semantic image retrieval~\cite{long2018towards}, and more recently recommender systems for new users~\cite{li2019zero}. Transfer learning seeks to improve a learner from one domain by transferring information from a related domain. Our proposed framework in this paper is a case of heterogeneous transfer learning~\cite{moon2017completely}, as the input spaces of the auxiliary and target tasks are different (See Figure~\ref{illustration}).

\textbf{Cold-start problem.} This mostly refers to modeling new users (no previous interactions with the system) or new items (no records of being consumed by users) in the application of recommender systems. Most methods assume some side feature information of the user or item is available, so that the representation of a new user or new item can be inferred. Such methods include matrix factorization~\cite{gantner2010learning, zhou2011functional}, pairwise regression~\cite{park2009pairwise}, decision trees~\cite{sun2013learning} and recently the zero-shot learning approach that uses linear encode and decode structure~\cite{li2019zero}.


\textbf{Semantic search.} It seeks to retrieve relevant items beyond keyword matching. The early effort of Latent Semantic Analysis (LSA)~\cite{deerwester1990indexing} uses a vector space to represent queries and documents.
More recently, neural IR models~\cite{mitra2018introduction,mitra2017learning,dai2018convolutional} seek to apply deep learning techniques for building (query, item) scoring models.
Supervised models~\cite{mitra2017learning,dai2018convolutional} are trained with search logs of (query, item) pairs. In contrast, unsupervised models~\cite{gysel2018neural} mostly learn the word and item representations purely based on the item's text features.
Recent work~\cite{zamani2020learning} shares the same motivation that search retrieval can learn from recommender system data. 
We point out several key differences between our paper and theirs. First being the different data assumptions, as their model is built on the (user, item) data with user embedding optimization. Our work does not require any explicit user data, and therefore, can fit to a wider range of applications. Second being the framework differences, as the recommender system data is not necessarily used for prediction target in our framework. Finally, our study focuses on real world search and recommender system with live experiment, which is not covered by their work.

\section{Problem Statement}
Denote a set of items $\mathcal{Y}=\{y_1,y_2,\cdots, y_n\}$ as the corpus of
a search and recommender system. Each item $y_i$ has text
feature $(x^{(i)}_1, x^{(i)}_2, \cdots, x^{(i)}_{k_i})$, which are from a vocabulary of $m$ words: $\mathcal{X} = \{x_1, x_2, \cdots, x_m\}$. The size $k_i$ of the text feature can vary for different item. Also the text feature can be either ordered (i.e., sequence) or unordered (i.e., set).
Finally, we use a
binary matrix $\mM \in \mathbb{R}^{n\times n}$ to represent
 (item, item) correlations, where $M_{i, j} = 1$ denotes item
$y_j$ is a neighbor of $y_i$: $j \in \text{Ne}(i)$.
In this paper, we do not require $\mM$ to be symmetric. For convenience, we also call the neighbor items
as context items. When there is no confusion, we use the terms embedding, vector, and representation interchangeably. 

Examples of the above problem setting include paper bibliography system, where items are individual papers, with their
text features coming from titles, abstract and keywords. The (item, item) correlations can be  derived by the citation network,
where two papers are correlated if one cites the other. Similarly in an online
music streaming system, the correlation matrix $\mM$ can
come from the music co-listening pairs, and the text features are the title, description or tags of the
music.

The task is to retrieve relevant items
given a query represented by a sequence of words $x_{e_1}, x_{e_2}, \cdots ,x_{e_p}$.
This would be a traditional supervised learning task if the training data (query, item) were provided.
However this data may not be available for newly built systems with limited user search activities. Or the data can be very sparse compared to the total number of queries and items.
On the other hand, the (item, item) correlation data commonly exists in most systems~\cite{hu2008collaborative}.
Therefore, we propose to encode the query and item to the same latent vector space by utilizing such (item, item) correlations. Then the search retrieval problem becomes nearest neighbor search~\cite{liu2005investigation}.
Formally,
a model outputs the latent space representation of words $\mW = [\vw_1, \vw_2,
\cdots, \vw_m] \in \mathbb{R}^{m\times d}$ and items $\mV = [\vec{v}_1, \vec{v}_2,
\cdots, \vec{v}_n] \in \mathbb{R}^{n\times d}$. The query embedding is:
\begin{equation}
    \vq = \text{encoder}(\vw_{e_1}, \vw_{e_2}, \cdots ,\vw_{e_p}),
    \label{eq:query}
\end{equation}
where the choice encoder can be (but not limited to) Bag-of-Words (BOW)~\cite{mikolov2013distributed}, which computes the mean of the word vectors, or (RNN)~\cite{cho2014learning}, or self-attention~\cite{vaswani2017attention}, which models the sequential relations of the words.
Then the top-$K$ candidate items are the ones with largest scores to the query:
$
\text{score}(\vq, \vec{v})
$,
where the score function can be either vector dot product or cosine
similarity.

\section{Zero-Shot Heterogeneous Transfer Learning Framework}
Before we introduce our proposed Zero-shot Heterogeneous Transfer Learning methods, we note that it is also possible to learn the semantic vector space of
words and items by using only the item text features. The idea is to use the item's text
feature as a proxy to the query. We can treat this learning task as a multi-class
classification task. For instance, when using softmax to represent the probability of item
$y_i$ given the text feature $x_{e_1}, \cdots, x_{e_k}$:
\[
\text{Pr}(y_i | x_{e_1}, \cdots, x_{e_k}) =
\frac{\text{exp}\big(\vec{v}_i^T \cdot \vq\big)}{\sum_{\ell}\text{exp}\big( \vec{v}_\ell^T \cdot \vq \big)},
\]
where $\vq = \text{encode}(\vw_{e_1}, \cdots, \vw_{e_k})$.
In this case, the total amount of training data equals the total number of items. 

Consider the special case of a BOW encoder. This method generalizes beyond
keyword matching by ensuring that the words with similar co-occurrence patterns will
be close in the semantic space. For instance, the words "obama" and "president" are likely to co-exist in
the same document. As a result, 
the optimization algorithm will not differentiate these two words much, so that they have similar word vectors.
Similarly, items with similar text features will be encoded closely in
the semantic space.

However, there are many cases that the words with similar semantic meanings do
not co-occur very often. For instance, if
the items rarely have both "funny" and "prank" as their text features, then the
learned vectors of these two words will be very different, and therefore, by
searching the query "funny", it is unlikely to retrieve the "prank" items. 
Fortunately, it is often common for pairs of items to form a connection in search and recommender system. For the above example, if users find these two kinds of items similar, they may link them as correlated items implicitly via interacting with the search and recommender system.

We propose to improve the target task of search retrieval, by transferring knowledge from (item, item) correlation, so that the semantic links between related words and items can be discovered. 

\begin{figure*}[!t]
\includegraphics[width=0.82\textwidth]{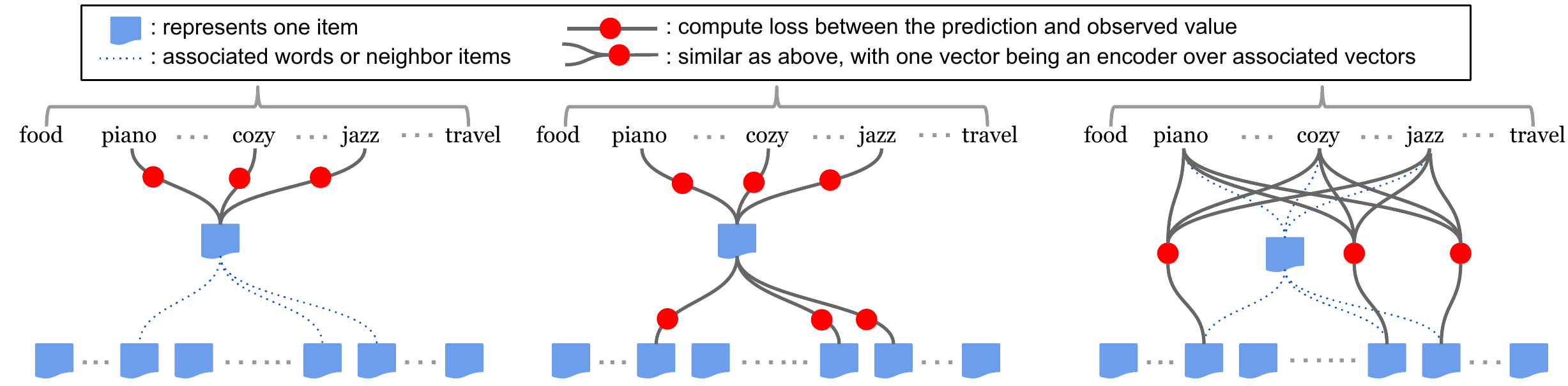}
\vspace{-0.1cm}
\caption{Illustration on the differences of the three models: Baseline Single Task Learning (STL) (\textbf{left}) that only trains Task 1 of ZSL\_ME. Proposed ZSL\_ME (\textbf{middle}) that jointly optimize the two tasks. Proposed ZSL\_TE (\textbf{right}) that optimize over (item, item) correlation.}
\label{fig:compare}
\vspace{-0.1cm}
\end{figure*}

\subsection{Zero-shot Learning: Multi-task Item Encoder}
\label{sec:multi_task}
Our first proposed way of transfer learning from (item, item) correlation matrix is to jointly optimize for the following two tasks:
\begin{itemize}
    \item Task 1: $\text{Text features of the item} \quad \underrightarrow{predict} \quad  \text{item}$.
    \item Task 2: $\text{Neighbor items of the item} \quad \underrightarrow{predict} \quad \text{item}$.
\end{itemize}

The specific model depends on the choice of prediction function and loss function. For instance, cross-entropy loss on a softmax prediction; square loss on a linear dot product prediction; pairwise ranking loss~\cite{wauthier2013efficient} with negative sampling. Here we only present the formulations with cross-entropy and square loss for simplicity.

Formally, the probability of predicting item $y_i$ given text feature encoder $\vq$ or given an item $y_j$ is as follows:
\[
\text{Pr}(y_i | \vq) =
\frac{\text{exp}\big(\vec{v}_i^T \cdot \vq\big)}{\sum_{\ell}\text{exp}\big( \vec{v}_\ell^T \cdot \vq \big)}
;\quad
\text{Pr}(y_i | y_j) =
\frac{\text{exp}\big(\vec{v}_i^T \cdot \vu_{j}\big)}{\sum_{\ell}\text{exp}\big( \vec{v}_\ell^T \cdot \vu_{j} \big)},
\]
where $\vq_i = \text{encoder}(\vw_{e_1}, \vw_{e_2}, \cdots ,\vw_{e_p})$ is the encoder of the text feature of $y_i$ and $\vu_j$ are the context vectors of $y_j$. The goal is to jointly optimize the cross-entropy (CE) loss for the above two tasks:
\[
\mathcal{L}_{CE} = - \sum_{i=1}^{n}\Big( \underbrace{\text{log} \big(\text{Pr}(y_i | \vq_i)\big)}_{\text{Task 1}} + \sum_{j\in \text{Ne(i)}} \underbrace{\text{log} \big( \text{Pr} (y_i | y_j) \big)}_{\text{Task 2}} \Big),
\]

Alternatively when modeling them as regression problem, we can use the following weighted square loss (SL):
\vspace{-0.1cm}
\begin{equation*}
\begin{split}
    \mathcal{L}_{SL} = & \sum_{i=1}^{n} \Big(\underbrace{(\vec{v}_i^T \cdot \vq_i - 1)^2 + \omega_0 \sum_{\ell \neq i} (\vec{v}_i^T \cdot \vq_\ell)^2}_{\text{Task 1}} \\
     + & \underbrace{\sum_{j\in\text{Ne}(i)}(\vec{v}_i^T \cdot \vu_j - 1)^2 + \omega_0 \sum_{\ell \notin \text{NE}(i) }(\vec{v}_i^T \cdot \vu_\ell)^2}_{\text{Task 2}} \Big),
\end{split}
\end{equation*}
where $\omega_0 < 1.0$ is the weight for implicit observations~\cite{hu2008collaborative}.


The optimization output are word vectors $\vw_i, 1\leq i \leq m$, item vectors $\vec{v}_i, q\leq i \leq n$ and context item vectors $\vu_i, 1\leq i \leq n$. The context vectors are akin to that in the language modeling (see word2vec~\cite{mikolov2013distributed}).
The reason we introduce context vectors instead of using the same item vectors is that, we want to encode two items to a close semantic space, if their neighbors are largely overlapped. By contrast, if we eliminate context vectors all together, and use item vectors for both prediction input and target: $\text{Pr}(y_i | y_j) = \text{exp}(\vec{v}_i^T \cdot \vec{v}_j)/\sum_\ell\text{exp}(\vec{v}_{\ell}^T \cdot \vec{v}_j)$, this effectively encodes the item close to its neighbors. From a matrix factorization perspective, eliminating context vectors is similar to symmetric matrix factorization on $\mM$ while $\mM$ is not necessarily symmetric.

We call this framework Zero-shot Learning with Multi-task Item Encoder (\textbf{ZSL\_ME}), as items are the shared prediction targets for both tasks.
The item vectors $\vec{v}_i$
serve the role of bridging the two optimization tasks. 
Specifically, vectors $\vec{v}_i$ and $\vec{v}_j$ will be close in the semantic space if their corresponding correlation patterns (i.e., $i-$th and $j-$th row of $\mM$) are similar.
Word vectors on the other hand are indirectly affected to encode such correlation information from $\mM$. Consider our previously example of words "funny" and "prank", where they do not co-occur often as the text features from the same item. However if their associated items share similar correlation pattern (i.e., similar neighbors of the items), those items will have similar embedding vectors (due to optimization on Task 2), and therefore the embedding vectors for these two words will also be close (due to optimization on Task 1).

\subsection{Zero-shot Learning: Item-to-Text Transformed Encoder}
\label{sec:transform}
Intuitively, there are two types of data relations, one is between the item and its text features, and other is between the item and its neighbor items (i.e., context items). The above ZSL\_ME method treats both relations as prediction targets.
In this section, we propose to model the (item, item) correlation as the only prediction task, and utilize the text features to encode context items.

Given the text feature $(x_{e_1}, x_{e_2}, \cdots, x_{e_p})$ for an item,  the context vector of this item is defined by
\begin{equation}
    \label{eq:encode_context}
    \vu = \text{encoder}(\vw_{e_1}, \vw_{e_2}, \cdots ,\vw_{e_p}).
\end{equation}
So the cross-entropy loss can be computed as:
\[
\mathcal{L}_{CE} = - \sum_{i=1}^{n}\sum_{j\in \text{Ne(i)}} \text{log} \big( \text{Pr} (y_i | y_j) \big)
;\quad
\text{Pr}(y_i | y_j) =
\frac{\text{exp}\big(\vec{v}_i^T \cdot \vu_{j}\big)}{\sum_{\ell}\text{exp}\big( \vec{v}_\ell^T \cdot \vu_{j} \big)}.
\]
And similarly the weighted square loss is:
\[
\mathcal{L}_{SL} = \sum_{i=1}^{n} \Big(
     \sum_{j\in\text{Ne}(i)}(\vec{v}_i^T \cdot \vu_j - 1)^2 + \omega_0 \sum_{\ell \notin \text{NE}(i) }(\vec{v}_i^T \cdot \vu_\ell)^2 \Big).
\]
Since the context vectors are represented by word vectors, the overall output from the above model are just word vectors $\vw_i, 1 \leq i \leq m$ and item vectors $\vec{v}_i, 1\leq i\leq n$. We call this framework Zero-shot Learning with Item-to-Text Transformed Encoder (\textbf{ZSL\_TE}), as the optimization on the correlation matrix trains the item and context item representations, which pass to the text representation due to a transformed encoder on text.

Here we explain our intuition of this proposed method.
Text features of a item are usually derived directly from the title or description of the item, which are precise information about the item. In this case, it is well fit to use them as item encoder instead of prediction target. On the other hand, the correlation matrix $\mM$ represents how much two items are similar or related content-wise. This relation usually extends beyond the text feature similarities. Therefore they are useful to be the prediction target. To demonstrate how this method generalizes to semantic related words that do not co-occur often, consider the same example of "funny" and "prank".
If their associated items (denote $y_r$ and $y_s$ correspondingly) co-occur as the neighbors of a same item $y_p$, then both $\vu_r$ and $\vu_s$ will be brought closer from the terms $\vec{v}_p^T \cdot \vu_r$ and $\vec{v}_p^T \cdot \vu_r$ in the loss functions. Since $\vu_r$ and $\vu_s$ are encoded by term "funny" and "prank" correspondingly, these two words end up close in the semantic space. Figure~\ref{fig:compare} illustrates how each model is different.

\subsection{Model Discussion}
\label{sec:discussion}
It is worth mentioning that Neural Collaborative Filtering (NCF)~\cite{he2017neural} method applies multi-layer perception instead of dot-product as an embedding combiner. We do not compare these two methodologies, as the choice of embedding combiner is not the central importance of this paper. We refer the interested readers to ~\cite{rendle2020neural} for more details. 

In the following, we discuss more details on the model options, training algorithms and practical usage examples.

\textbf{Regularization.} To prevent model overfitting, the regularization term $\lambda ||\bm{\theta}||^2_{F}$ can be added to the model loss functions. The above $\lambda$ is the co-efficient for the regularzation term, and the symbol $||\cdot||^2_{F}$ denotes the matrix Frobenius norm, and $\bm{\theta}$ denotes the matrix of all parameters. $\bm{\theta} = [\vw_1, \cdots, \vw_m, \vu_1, \cdots, \vu_n, \vec{v}_1, \cdots, \vec{v}_n]$ for ZSL\_ME and $\bm{\theta} = [\vw_1, \cdots, \vw_m, \vec{v}_1, \cdots, \vec{v}_n]$ for ZSL\_TE.

\textbf{Weighted Training.} Real word correlation data are usually skewed towards popular items (a.k.a. power law)~\cite{faloutsos1999power}. In other words, the number of non-zeros (i.e., nnz) of the rows (or columns) in $\mM$ can be concentrated in only a few rows (or columns). We can reweight the training example to avoid having the loss function dominated by only a few most popular items:
\[
\sum_{i=1}^{n}\sum_{j\in \text{Ne}(i)} \text{loss}(y_i, y_j)\quad \underrightarrow{\text{reweight}} \quad \sum_{i=1}^{n}\sum_{j\in \text{Ne}(i)} r_i c_j \text{loss}(y_i, y_j),
\]
where $r_i$ are row weights and  $c_j$ are column weights. In this paper, we set them to be proportional to $1/\sqrt{nnz}$ of the corresponding rows and columns, and rescale them so that the mean values of both row weights and column weights are $1.0$.

\textbf{Choice of Encoder.} The encoder is applied on the text features of an item during training, as well as on the query words during serving. Depending on the text feature characteristics (e.g., ordered vs unordered, long vs short) and the specific application domain, various encoders can be used. Bag-of-Word (BOW) is simplest to use as it doesn't have underlying requirement on the text feature format. RNN can be used to encode a sequence of words. And self-attention mechanism can usually work well for long sentence or document~\cite{cho2014learning}. In this paper, we use BOW in our experiments for simplicity and also for reducing the complexity in order to work with live experiment serving constraints. 

\textbf{Loss Function \& Optimization.}
For large scale search and recommender systems, the number of items can be millions or beyond. This imposes computational challenges when computing the softmax function for $\mathcal{L}_{CE}$ or iterating through all negative examples for $\mathcal{L}_{SL}$, which yields $O(n^2)$ complexity. Approximation solutions include sampled softmax, hierachical softmax and negative sampling. They share the core idea of avoiding to exhaustively iterate through all items, so that optimization like stochastic gradient descent (SGD) can be applied. In a special case using BOW encoder with square loss $\mathcal{L}_{SL}$,
the first-order and second-order derivative of the full loss function can be efficiently computed, without the need of explicitly iterating through $O(n^2)$ negative examples~\cite{bayer2017generic}. Therefore the optimization problem can be solved by the coordinate descent algorithm. With this advantage of computing the full negative examples without the need to do negative sampling, we therefore choose to use $\mathcal{L}_{SL}$ instead of $\mathcal{L}_{CE}$ in our experiments.

\section{Experiments}
We first conduct a set of offline evaluations (Sections \ref{sec:correlation_matrix}, \ref{sec:human}, \ref{sec:search}) among our proposed methods ZSL\_ME and ZSL\_TE and baseline methods. Then we conduct the  live experiment (Section \ref{sec:live}) with the best ZSL method on one of the world’s largest search and recommender systems from Google.


\textbf{Dataset.} Each item $y_i$ is a product of the recommender system. We derive the correlation matrix $\mM$ from sequential item consumptions. Formally, if $y_p$ is consumed right after $y_q$ by the same user, then $y_p$ is the neighbor of $y_q$ (i.e., $M_{q, p} = 1$). To reduce the noise of the correlation matrix, we rank each seed item's neighbors by the counts of their co-occurrences with the seed item, and only keep the top $250$ of them. So each row of $\mM$ has at most $250$ non-zeros. Titles and descriptions are used as the text features of each item. The word vocabulary contains both unigrams and bigrams from those text features. We threshold the minimal occurrences for items and words, and the final vocabulary size is $17,714,821$ for items, $2,451,962$ for words. 
The average number of neighbors for an item is $176$, and the average number of words of an item is $155$.


\textbf{Experiment Settings.} One baseline model is the single task learning version of ZSL\_ME, where it only trains on Task 1 that predicts the item given its text features. We call it Single Task Learning (\textbf{STL}).
It is important to note that, STL is proven to be a more effective approach than traditional methods like LSI in many of the recommendation tasks~\cite{hu2008collaborative,bobadilla2013recommender} as it models implicit feedbacks. 
In addition we also trained a supervised multiclass classification model directly based on the actual search data. Each record (query, item) corresponds to one item consumption from the search query. Formally, the word and item representations are trained to minimize the cross entropy loss of the sampled softmax. Although this supervised method is outside of the zero-shot framework of this paper, it is interesting to compare them and study how they are different. Here are the detailed setting for each method:
\begin{itemize}
    \item \textbf{STL}: single-task square loss $\mathcal{L}_{SL}$ without combining (i.e., encoder) text feature; $\omega_0 = 0.001, \lambda = 4.0$; trained with $10$ iterations of coordinate descent. 
    \item \textbf{ZSL\_ME}: same setting as above multi-task square loss.
    \item \textbf{ZSL\_TE}: square loss $\mathcal{L}_{SL}$ with the BOW encoder; same setting as the above two.
    \item \textbf{Supervised Multiclass Classification (SMC)}: BOW encoder on words; trained with a total of $1780$ millions (query, item) pairs; sampled softmax with $10000$ negative samples for each batch with batch size $128$; trained with SGD optimizer with learning rate $0.06$ for $50$ million iterations.
\end{itemize}
The above hyperparameters are set based on empirical hyperparameter search. For all methods, the embedding dimension is set as $200$. The following sections include the semantic retrieval task: given a seed vector $\vq$ (e.g., query or item), the task is to retrieve top-$K$ items $\vec{v}$ with the highest $\text{score}(\vq, \vec{v})$. By default (unless otherwise stated) we use cosine score for STL, ZSL\_ME and ZSL\_TE, and vectors dot product for SMC\footnote{We found dot product works much better than cosine for SMC for all tasks. More discussions in Section~\ref{sec:search}.}.

\begin{table}[!tb]
\centering
\begin{tabular}{c cccc}
\toprule
 & STL & SMC & ZSL\_ME & ZSL\_TE \\
\cmidrule{2-5}
Recall (\%) & $17.8$ & $13.8$ & $27.4$ & $\bm{35.0}$ \\
\bottomrule
\end{tabular}
\caption{Recalls of the correlation matrix reconstruction task. Note a random guess would only yield $0.001\%$ recall in finding candidates from a pool over $17$ million items.}
\label{tab:reconstruct}
\vspace{-0.5cm}
\end{table}
\subsection{Correlation Matrix Reconstruction Task}

\label{sec:correlation_matrix}
First we evaluate how each method performs in terms of reconstructing the correlation matrix $\mM$. We use recall to evaluate percentage of relevant items being retrieved at top positions. This is commonly used for retrieval and top-n item recommendation models. Specifically, for each item $y_i \in \mathcal{Y}$, denote $\mathcal{S}_{true, i}$ as the non-zeros in the $i-$th row of the correlation matrix $\mM$ and $k_i$ as the size of $\mathcal{S}_{true, i}$ .
Denote $\mathcal{S}_{pred, i}$ as the retrieved top-$k_i$ items $\vec{v}_\ell$ with the highest $\text{score}(\vec{v}_i, \vec{v}_\ell)$, the recall for item $y_i$ is defined as:
\[
\text{recall}_i = \frac{|\mathcal{S}_{true, i} \cap \mathcal{S}_{pred, i}|}{|\mathcal{S}_{true, i}|},
\]
then we report the average recall over all items. 
We believe using recall for evaluation is more intuitive than directly comparing the square losses of different methods, because the square loss greatly depends on the hyperparameters, and the value itself does not reflect any application meaning.



\textbf{Results.} Table~\ref{tab:reconstruct} shows the recalls of all methods. We find the two proposed transfer learning methods ZSL\_ME and ZSL\_TE outperform STL and SMC. This is expected, as STL and SMC are trained without any information of the (item, item) correlation data. This observation shows that the outputs from our proposed framework are indeed influenced by the recommender system data in a positive way. Finally, we notice that between the two proposed methods, ZSL\_TE is superior than ZSL\_ME. We also observe this pattern for the other tasks as well in this paper. Our hypothesis is that directly optimizing two tasks in ZSL\_ME could potentially introduce conflicts~\cite{sener2018multi}, while ZSL\_TE does not have this issue.

\subsection{Offline Retrieval Task on Human Labeled Data}
\label{sec:human}
In this section, we evaluate how relevant are the retrieval results when using human labeled (query, item) ground truth pairs.
We have a list of queries and their corresponding items that are labeled as relevant by human. We then form a candidate pool as the union of all relevant items from these queries, and a target set for each query as its relevant items.

\begin{table}[!tb]

\centering
\begin{tabular}{c cccc}
\toprule
 & \#query & $\mathcal{S}$ & average $|\mathcal{S}_i|$ & $\sum_{i=1}^{r}|\mathcal{S}_i| / |\mathcal{S}|$ \\
\cmidrule{2-5}
Eval Set 1 & $202$ & $23,377$ & $136.5$ & $1.18$ \\
Eval Set 2 & $125$ & $15,842$ & $174.5$ & $1.38$ \\
Eval Set 3 & $318$ & $40,267$ & $144.7$ & $1.14$ \\
Eval Set 4 & $36$ & $1,797$ & $102.1$ & $2.05$ \\
\bottomrule
\end{tabular}
\caption{Statistics of the evaluation set from human labeled data. Each column represents: the total number of the queries; size of the candidate pool (i.e., the joint of target sets of all queries); average size of target set; average number of appearance for items.}
\label{tab:evaldata}
\vspace{-0.4cm}
\end{table}

Formally, consider $r$ different queries, and each query corresponds to a set of relevant items denoted as $\mathcal{S}_1 = \{y^{(1)}_1, y^{(1)}_2, \cdots, y^{(1)}_{k_1}\}$, $\mathcal{S}_2 = \{y^{(2)}_1, y^{(2)}_2, \cdots, y^{(2)}_{k_2}\}, \cdots, \mathcal{S}_r = \{y^{(r)}_1, y^{(r)}_2, \cdots, y^{(r)}_{k_r}\}$. We can denote their joint set as $\mathcal{S} = \mathcal{S}_1 \cup \mathcal{S}_2 \cup \cdots \cup \mathcal{S}_r$. Given a query (for instance the $i-$th one), we select $k_i$ items from $\mathcal{S}$ with the highest score to the query, and denote it as $\mathcal{Z}_i = \{\tilde{y}^{(i)}_1, \tilde{y}^{(i)}_2, \cdots, \tilde{y}^{(i)}_{k_i}\}$. Then the recall for this query is defined as:
\[
\text{recall}_i = \frac{|\mathcal{S}_i \cap \mathcal{Z}_i|}{|\mathcal{S}_i|}.
\]
Intuitively, the above recall denotes the percentage of relevant $k$ items are among the top-$k$ predictions.

We note that it is possible for certain items to exist in multiple the ground truth set $\mathcal{S}_i, i=1, \cdots, r$ (i.e., they are not disjoint). This actually makes the task more difficult, because we select the group of queries to share some common attributes, so that for each query, the unrelated items (i.e., related to other queries) are not too distinct. We can define the average number of appearance for items as $\sum_{i=1}^{r}|\mathcal{S}_i| / |\mathcal{S}|$ in order to show the degree of overlapping on these sets. We have four such datasets, with each representing one category of the queries. For instance, one dataset has all its queries and items being music related. See Table~\ref{tab:evaldata} for their statistics.

\begin{figure}[!tb]
  \begin{center}
    \includegraphics[width=0.4\textwidth]{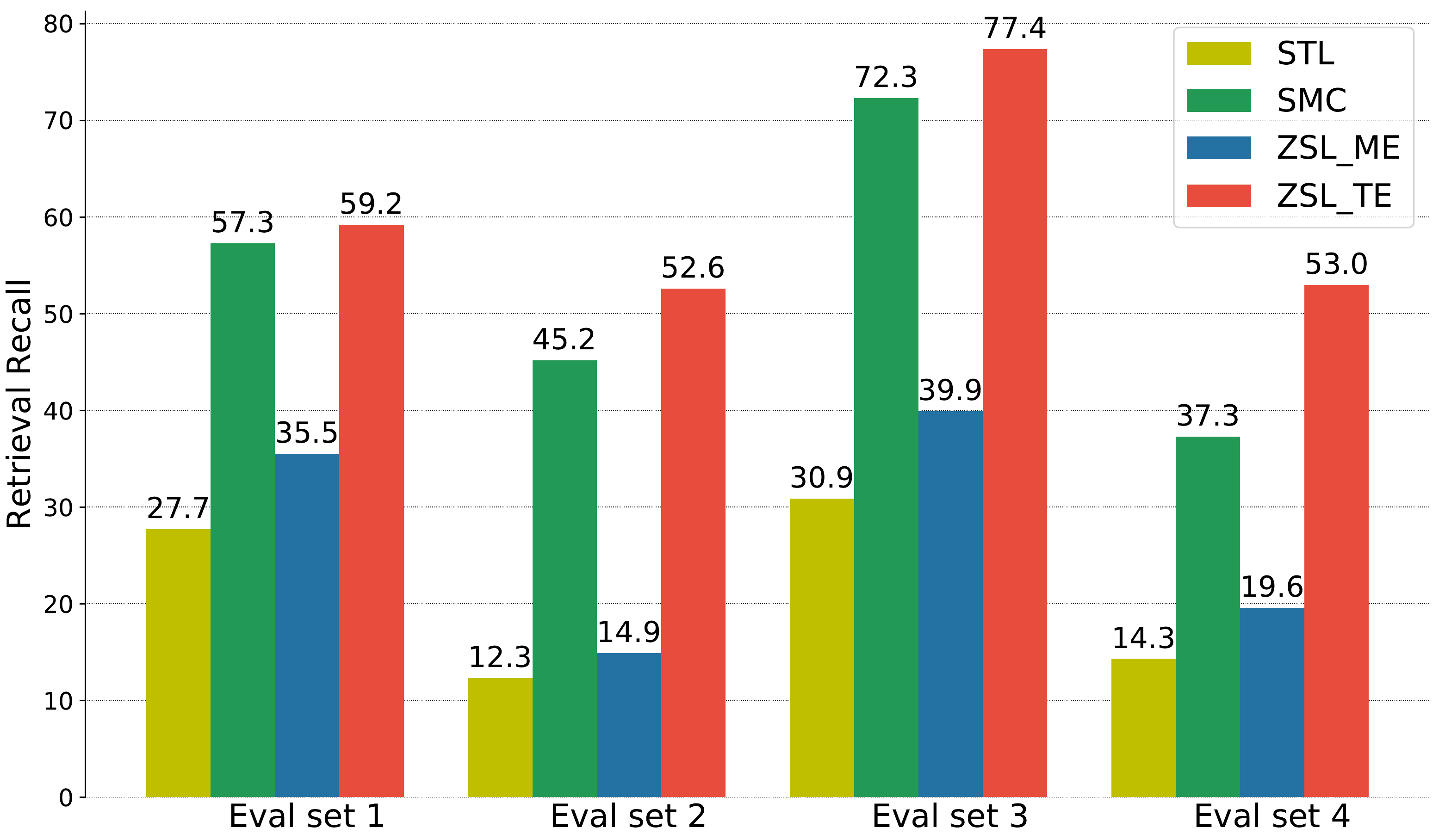}
  \end{center}
  \vspace{-0.25cm}
  \caption{Comparison of different methods on the recall for the offline retrieval task on human labeled data. For each evaluation set, the average recall of all queries in this set is reported.}
  \vspace{-0.5cm}
  \label{fig:retrieval_accuracy}
\end{figure}

\textbf{Results.} Figure~\ref{fig:retrieval_accuracy} shows the comparison results for all methods on this task.
We can see that our proposed transfer learning approaches ZSL\_ME and ZSL\_TE outperforms STL. This shows incorporating the (item, item) information from the recommender system can improve the semantic search retrieval task. Consistently, between these two transfer learning approaches, ZSL\_TE performances much better.
We also notice that ZSL\_TE even achieve higher recall than the supervised method SMC. The explanation is that, although the supervised method is directly trained on (query, item) data, but the evaluation sets are extracted from human labeled data, which encodes only the relevance information between query and item. By comparison, the actual user search data (query, item) is affected by popularity of the item or content of the item. Because for a large scale recommender system, hundreds of thousands of items can be relevant to a query, but only those that are appealing enough will be clicked by users. We will later discuss evaluation results in Section~\ref{sec:search} on the actual search data. 


\subsection{Offline Retrieval Task on Search Data}
\label{sec:search}
In this section, we use the ground-truth pairs (query, item) from the search logs. Note that this is the same data source as the training data of SMC.
We hold out $1$ million such pairs for evaluation. Different from the human labeled data, the search data reflects not only relevance, but also users' preferences. For large-scale search and recommender system, certain items can be very relevant to a query, but may not necessarily be preferred by users.

We use the metric $\text{recall}@K$ for evaluation, which is defined as the ratio of the ground-truth items at the top-$K$ retrieved list of the method.

\begin{figure}[!tb]
  \begin{center}
    \includegraphics[width=0.35\textwidth]{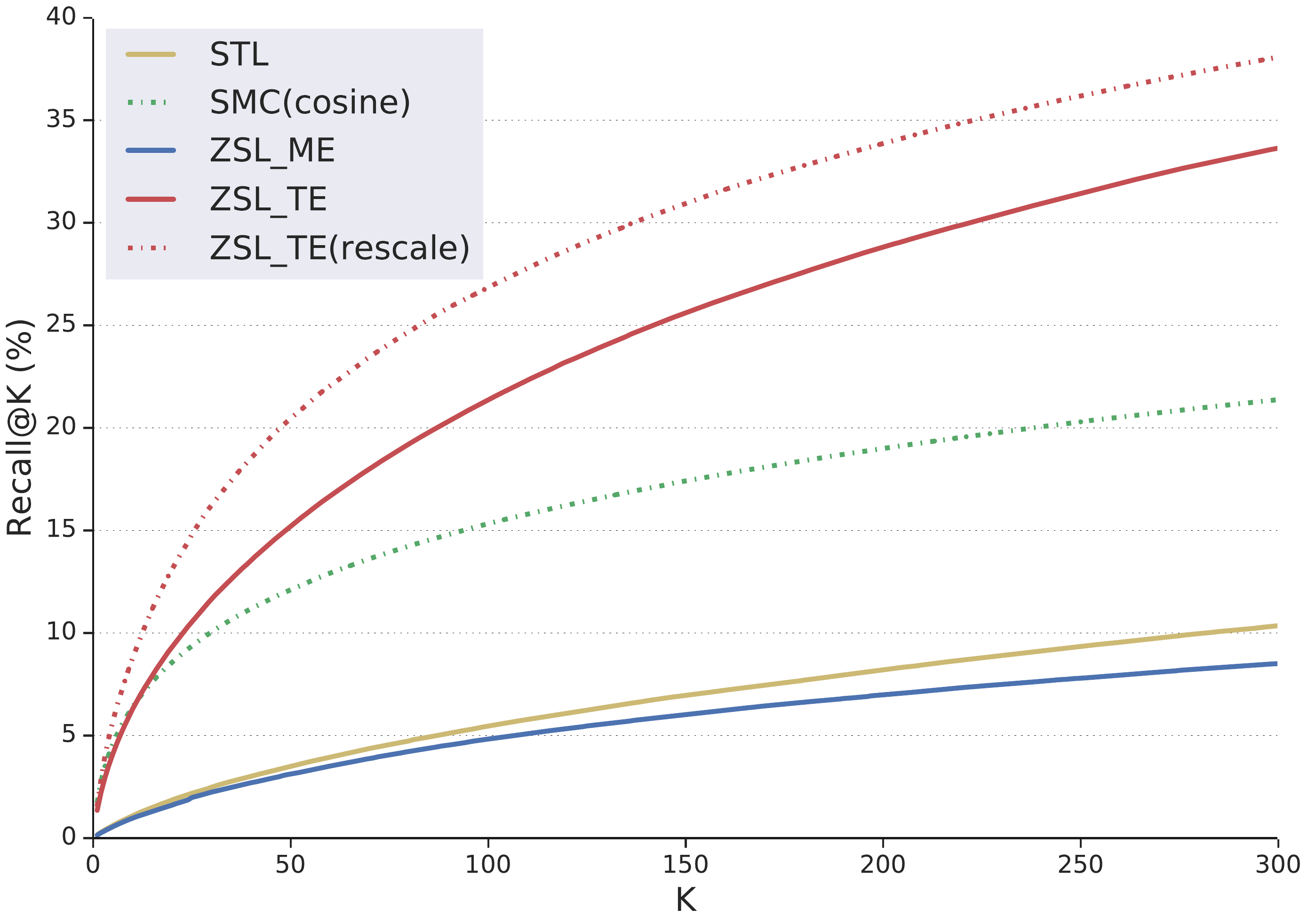}
  \end{center}
  \vspace{-0.25cm}
  \caption{Recall@K for the evaluation task on search data. Dotted lines are two added methods used for demonstration purpose. SMC(cosine) is defined as using the SMC method for retrieval with cosine similarly instead of dot product as score function. ZSL\_TE(rescale) is defined as rescaling the item vectors of ZSL\_TE method by borrowing the norms of item vectors learned in SMC, and the retrieval score function is dot product.}
  \label{fig:search_retrieval_accuracy}
\end{figure}

\textbf{Results.} Figure~\ref{fig:search_retrieval_accuracy} shows the comparison results of different methods for this task.
SMC is not presented in the figure as it is the supervised method that is trained on the exact same data source of the evaluation task, so it is expected to outperforms all the unsupervised methods by large margins. In our case, SMC could reach $73.6\%$ for $\text{recall}@300$.
We can clearly notice that our proposed ZSL\_TE achieves superior recalls compared to ZSL\_ME and STL. So far, all the offline evaluations show that ZSL\_TE performs consistently better than ZSL\_ME, and in this task, ZSL\_ME is even slightly worse than STL. Our hypothesis is the same as stated in section~\ref{sec:correlation_matrix} that multitask optimization could introduce additional conflicts, which is commonly observed in related researches~\cite{sener2018multi}. We also notice that, if we change the SMC retrieval method from vector dot product to cosine (as is used by all other methods), the recalls are beat by our ZSL\_TE. This provides us an interesting insight that the vector norms in the SMC method yield useful information about user preferences over items. To verify this hypothesis, we rescaled the item vectors of ZSL\_TE based on the norms of SMC, and find out the recall is improved.\footnote{The same observation holds also for ZSL\_ME and STL. They are skipped in Figure~\ref{fig:search_retrieval_accuracy} for the sake of simplicity.}


In practice, our proposed methods are not meant to replace the existing supervised models, but to serve as additional sources of candidates generation.
Here we show that combining our method to SMC can improve the recall, even with the following trivial ensemble rule.
The new item list has the first $150$ items the same as SMC, then the next $150$ items are from ZSL\_TE and the rest of SMC in a interleaved way. Table~\ref{tab:ensemble} shows the comparison results. It is also important to note that the benefits of ensembling our ZSL\_TE with supervised methods can be more pronounced, because the current evaluation is done on the offline search data, which is already limited in scope as it does not answer the question of what if we recommend the user the other item instead of the current one. This effect can only be assessed from the live experiment (Section~\ref{sec:live}). 

\begin{table}[!tb]

\centering
\begin{tabular}{c p{0.09\linewidth}p{0.12\linewidth}p{0.14\linewidth}}
\toprule
 & \small{SMC} & \small{Ensemble} & \small{Ensemble (rescale)} \\
\cmidrule{2-4}
\small{Recall@300 (\%)} & $73.6$ & $75.4$ & $\bm{76.2}$\\
\bottomrule
\end{tabular}
\caption{Recall@300 for ensembling ZSL\_TE or ZSL\_LE (rescale) with SMC for the offline retrieval task on search data. Here the recall is improved even with the naive ensemble rule.}
\label{tab:ensemble}
\vspace{-0.3cm}
\end{table}


\subsection{Live Experiment}
\label{sec:live}
\textbf{Settings.} In this section, we evaluate ZSL\_TE in an A/B live experiment.
The control group is the production search system that is highly-optimized 
with many components of advanced retrieval techniques (e.g., both term-frequency-based scoring and supervised machine learning algorithms).
The experiment group is an ensemble of our ZSL\_TE and the production system. Each time when a query comes, the ZSL\_TE retrieves top-$100$ items, and the ensemble algorithm will jointly rank (i.e., based on query features as well as item features) these items with those retrieved by the control production system.

\textbf{Model Daily Refresh.} We do warm-start training with the new data everyday to include new items and words to the model. Specifically, we load the existing model and only need to conduct a few model training iterations to make the embeddings of new items and words converge.

\textbf{Evaluations.} We conduct this live experiment over several millions of queries. We compare the following evaluation metrics:
\begin{itemize}
    \item Query Coverage: The ratio of queries that receive at least one user interaction.
    \item User Interaction: The level of interactions between users and the items.
    \item Query Refinement: The proportion of queries that have a follow up query sharing a keyword. Smaller value is better (meaning users are more satisfied with the results).
    \item Next-page CTR: The proportion of queries that lead to a next page click by users. Smaller value is better (meaning users are satisfied with the first page of results).
    \item Human-rated Relevance Score: A numerical score representing the relevance of the retrieved items to a query, as evaluated by trained human raters. We randomly sample $250$ instances, each with a query and a list of retrieved items, for both the control group and the experiment group.
\end{itemize}
We also have several ranking metrics that are similar to the normalized discounted cumulative gain (nDCG), but they are tailored to our specific system, therefore less generic. We do not include them in this paper to avoid confusion. But it is important to mention that our live experiment also shows significant improvements on these metrics as well.

\begin{table}[!tb]
\centering
\begin{tabular}{ccc}
\toprule
 & A/B Test Diff & Notes \\
\cmidrule{2-3}
Query Coverage & $\bm{+0.51}\ (\pm0.27)\%$  \\
User Interaction & $\bm{+2.52}\ (\pm1.54)\%$ \\
Query Refinement & $\bm{-0.94}\ (\pm0.57)\%$  & lower is better\\
Next-page CTR & $\bm{-1.60}\ (\pm0.75)\%$ & lower is better \\
Human Rater Score & $\bm{+14.0} \%$ \\
\bottomrule
\end{tabular}
\caption{Live experiment metrics. The bold numbers are statistically significant, and the numbers after $\pm$ are the standard deviations.}
\label{tab:online}
\vspace{-0.5cm}
\end{table}

\textbf{Results.} As shown in Table~\ref{tab:online}, we observe significant improvements including increased query coverage, decreased query refinements (meaning users are more satisfied with the results), increased user interaction, and higher relevance scores from human raters. These results demonstrate that the proposed approach is effective on semantic search task, even without training on any search data.

\begin{figure}[!tb]
  \begin{center}
    \includegraphics[width=0.45\textwidth]{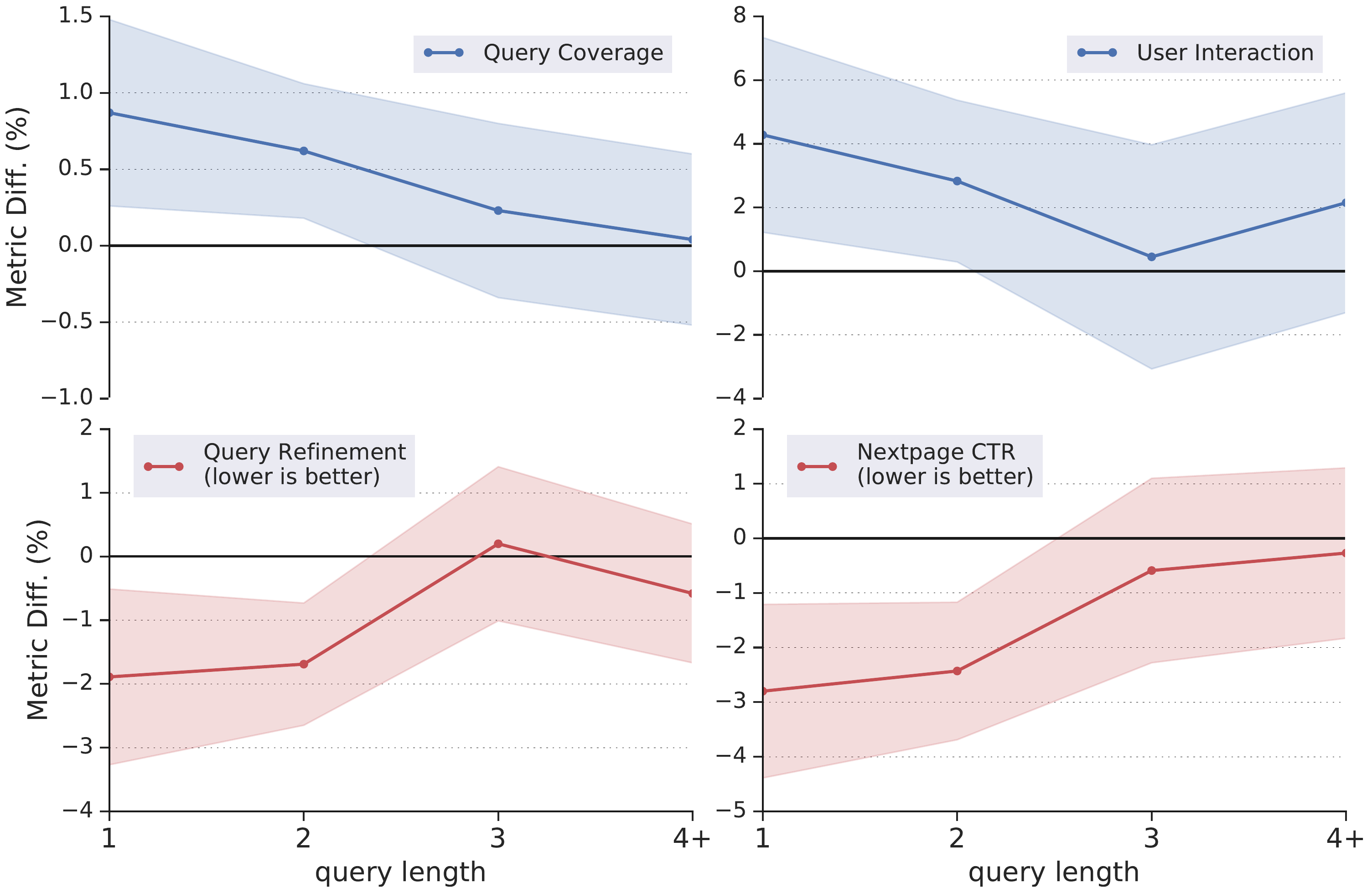}
  \end{center}
  \vspace{-0.1cm}
  \caption{Live experiment metrics break down by the length of query. The lines represent the mean values, and the colored fills are standard deviations. Two different colors to differentiate higher (blue) or lower (red) is better for the corresponding metrics.}
  \label{fig:live_split}
\vspace{-0.15cm}
\end{figure}

We also notice that the our method has a larger improvement when the query length (number of unigrams) is small (see Figure~\ref{fig:live_split}). Since shorter queries often correspond to broader user intent, there are usually more relevant items per query, which means more (query, item) pairs were previously unseen in the supervised training data. This is an evidence that, our zero-shot transfer learning framework can result in more improvement by solving the cold start problem.

\section{Conclusions and Lessons Learned}

In this paper, we explore a Zero-Shot Heterogeneous Transfer Learning framework, that trains a model to learn the natural-language and item representations based on the (item, item) correlations derived from the recommender system, and then uses the learned representations to serve the search retrieval task. 

Here are several lessons we learned. 1) Both the proposed ZSL\_ME and ZSL\_TE show evidences that it is beneficial for search retrieval system to transfer learn from the recommender system. However ZSL\_LE works much better with the real data in this paper. This is a valuable finding,
as it suggest that treating the text features as rules and predicting the (item, item) correlations works better in practice than predicting both correlations.
2) We find the effectiveness of the supervised method SMC is largely due to "memorization" based on our observation that the norms of embeddings play a crucial role. This is evidence that supervised methods such as SMC are less capable of generalizing beyond seen (query, item) pairs. 3) Most importantly, we find in live experiment, that our proposed ZSL\_TE are more effective for broad queries (as inferred by query length). This insight suggests the direction of investigating different query types, so that our model can be deployed in a most effective way.

\bibliographystyle{ACM-Reference-Format}


\end{document}